\documentclass[10pt,twocolumn,letterpaper]{article}
\usepackage[pagenumbers]{cvpr} % To force page numbers, e.g. for an arXiv version

\definecolor{cvprblue}{rgb}{0.21,0.49,0.74}

\usepackage[pagebackref,breaklinks,colorlinks,allcolors=cvprblue]{hyperref}
\usepackage{comment, enumitem, amsfonts, multirow, booktabs, array, arydshln, amssymb, csquotes,makecell,float}
\def\paperID{8677} % *** Enter the Paper ID here
\def\confName{CVPR}
\def\confYear{2025}

\title{Topological Symmetry Enhanced Graph Convolution for Skeleton-Based Action Recognition}

\author{Zeyu Liang \quad Hailun Xia\footnotemark \quad Naichuan Zheng \quad Huan Xu\\
 School of Information and Communication Engineering\\
Beijing University of Posts and Telecommunications\\
{\tt\small \{lzy\_sfading, xiahailun, 2022110134zhengnaichuan, xuhuan\}@bupt.edu.cn}
}

\begin{document}
\maketitle
\begin{abstract}
Skeleton-based action recognition has achieved remarkable performance with the development of graph convolutional networks (GCNs). However, most of these methods tend to construct complex topology learning mechanisms while neglecting the inherent symmetry of the human body. Additionally, the use of temporal convolutions with certain fixed receptive fields limits their capacity to effectively capture dependencies in time sequences. To address the issues, we (1) propose a novel Topological Symmetry Enhanced Graph Convolution (TSE-GC) to enable distinct topology learning across different channel partitions while incorporating topological symmetry awareness and (2) construct a Multi-Branch Deformable Temporal Convolution (MBDTC) for skeleton-based action recognition. The proposed TSE-GC emphasizes the inherent symmetry of the human body while enabling efficient learning of dynamic topologies. Meanwhile, the design of MBDTC introduces the concept of deformable modeling, leading to more flexible receptive fields and stronger modeling capacity of temporal dependencies. Combining TSE-GC with MBDTC, our final model, TSE-GCN, achieves competitive performance with fewer parameters compared with state-of-the-art methods on three large datasets, NTU RGB+D, NTU RGB+D 120, and NW-UCLA. On the cross-subject and cross-set evaluations of NTU RGB+D 120, the accuracies of our model reach 90.0\% and 91.1\%, with 1.1M parameters and 1.38 GFLOPS for one stream. Code will be available at \url{https://github.com/Sfadingz/TSE_GCN}.

% \footnotemark
% \footnotetext{Code is available at.}
\footnotetext{* \quad Corresponding author.}
\end{abstract}
    
\section{Introduction}
\label{sec:intro}

Human action recognition has attracted much attention due to its wide range of applications\cite{bates2017line, elkholy2019efficient, jiang2015human,poppe2010survey}, including video surveillance, virtual reality, health care and so on. With the development of depth sensors\cite{zhang2012microsoft, yeung2021effects} and human pose estimation methods\cite{sun2019deep,cheng2020higherhrnet,cao2017realtime}, skeleton-based action recognition\cite{kim2018disentangling,plizzari2021skeleton,chen2021channel,yan2018spatial} has become increasingly popular. Skeleton is a type of structured data with each joint of the human body identified by its joint type, frame index and 3D position, which shows great potential in preserving privacy and demonstrates strong robustness against the variations of illumination, viewpoints and other background changes. 

\begin{comment}
\begin{figure}[t]
\centering
\begin{subfigure}{0.8\linewidth}
%\fbox{\rule{0pt}{2in} \rule{.9\linewidth}{0pt}}
\includegraphics[width=1.0\linewidth]{images/TSEGC.pdf}
\caption{An example of a subfigure.}
\label{fig:short-a}
\end{subfigure}
\begin{subfigure}{0.8\linewidth}
% \fbox{\rule{0pt}{2in} \rule{.9\linewidth}{0pt}}
\includegraphics[width=1.0\linewidth]{images/temporalconv.pdf}
\caption{Another example}
\label{fig:short-b}
\end{subfigure}
\caption{Example of a short caption, which should be centered.}
\label{fig:short}
\end{figure}
\end{comment}

\begin{figure}[t]
	\centering
	%\fbox{\rule{0pt}{2in} \rule{0.9\linewidth}{0pt}}
	\includegraphics[width=1.0\linewidth]{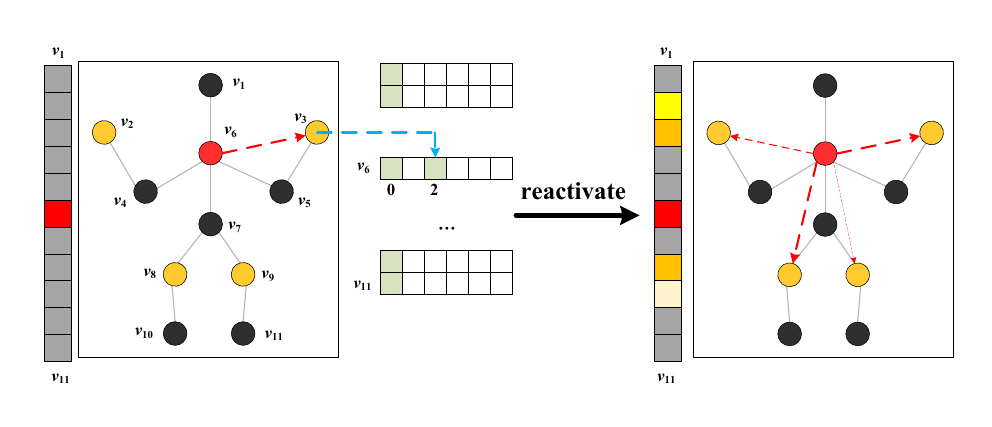}
	
	\caption{Topology reactivation with symmetry awareness. The correlations between \textit{$v_6$} and \textit{$v_2$}, \textit{$v_3$}, \textit{$v_8$}, \textit{$v_9$} in the shared topology are activated due to the scale mask derived from the correlation between \textit{$v_6$} and \textit{$v_3$}. Darker colors and thicker lines stand for larger weights. Best viewed in color.}
	\label{fig:mechanism}
\end{figure}

Inspired by the inherent structure of skeleton data, graph convolutional networks (GCNs) have emerged as a dominant solution for skeleton-based action recognition. STGCN\cite{yan2018spatial} was the first work to encode human body as spatial temporal graphs, aggregating features along the natural connectivity of the human body. Since then, many works\cite{shi2019two,lee2023hierarchically,chen2021channel,liu2020disentangling,shi2019skeleton,li2019actional} have delved into the optimization of graph convolutional mechanisms, with a focus on adaptively building topologies to effectively capture motion patterns. However, most of these approaches leverage learnable adjacency matrices combined with attention mechanisms and other techniques derived from dynamics theory\cite{wang2023neural}, information bottleneck theory\cite{chi2022infogcn}, topological data analysis\cite{zhou2024blockgcn} and so on, neglecting or underestimating the inherent symmetry of the human body and its potential role in topology learning. For instance, in the case of a \enquote{brush hair} action, the interaction between the head and the left hand should be hight-lighted and learned alongside the interaction between the head and the right hand, while the roles of other joints may be less critical.

Based on the analysis above, we argue that integrating topological symmetry constraints into topology learning could lead to a straightforward yet effective mechanism for topological learning. In this paper, we propose a topological symmetry enhanced graph convolution, named TSE-GC, to enable distinct topology learning across different channels groups while incorporating topological symmetry awareness. Specifically, TSE-GC learns the scale mask of interest for each sample and reactivates a shared topology into distinct copies for multiple separate channel groups (illustrated in Fig.\ref{fig:mechanism}). The scale mask is learned via k-nearest neighbor (k-NN) algorithm based on the Euclidean distance, where the k nearest neighbors of each joint and the corresponding scales are identified. During the learning of scale mask, we also introduce the calculated Euclidean distance as a calibration for the topology tailored to each sample. With few extra parameters introduced, our method facilitates topology learning in a symmetric manner and effectively captures the intricate correlations between joints.

In terms of feature aggregation across different frames, most of these approaches\cite{liu2020disentangling,chen2021channel,zhou2024blockgcn} tend to apply multi-scale temporal convolution to effectively capture both short-range and long-range dependencies, yet their representation capacity is still limited due to essentially fixed receptive fields of temporal convolution. To remedy the issue, we introduce the concept of deformable convolution and construct a multi-branch deformable temporal convolution, named MBDTC, for skeleton-based action recognition. MBDTC incorporates learnable offsets to the sampling locations of the temporal convolution filter, enabling more flexible receptive fields and better representation capacity. 

Combining TSE-GC with MBDTC, we construct a novel topological symmetry enhanced graph convolution network (TSE-GCN) for skeleton-based action recognition. Extensive experimental results on NTU RGB+D\cite{shahroudy2016ntu}, NTU RGB+D 120\cite{liu2019ntu}, and NW-UCLA\cite{wang2014cross} show that TSE-GCN achieves competitive performance with fewer parameters compared with state-of-the-art methods.

Our contributions can be summarized as follows:

\begin{enumerate}
\item We propose a topological symmetry enhanced graph convolution (TSE-GC) which facilitates topology learning by incorporating physical constraints rooted in topological symmetry and flexibility.

\vspace{0.5em}
\item We introduce the concept of deformable modeling and construct a multi-branch deformable temporal convolution (MBDTC) for skeleton-based action recognition with enhanced representation capacity.

\vspace{0.5em}
\item Integrating TSE-GC with MBDTC, Our TSE-GCN achieves competitive performance against state-of-the-art methods on three large datasets for skeleton-based action recognition with fewer parameters.
\end{enumerate}

\section{Related Work}
\label{sec:formatting}

%-------------------------------------------------------------------------
\subsection{Skeleton-based Action Recognition}

Early deep-learning methods utilized recurrent neural networks (RNNs)\cite{du2015hierarchical,wang2017modeling} and convolutional neural networks (CNNs)\cite{li2017skeleton,soo2017interpretable} to capture action representations by encoding skeleton data as feature sequences or pseudo-images, but they overlooked the inherent relations between joints. Consequently, these methods failed to effectively model human body topology, limiting their overall performance. In contrast, GCNs represent the human body as graphs, where joints are treated as nodes and their relations are treated as edges. Such a design enables GCNs to effectively capture dependencies between joints and brings the potential to unravel complex motion patterns.

\subsection{GCNs for Skeleton-based Action Recognition}

Topology learning is a key focus of GCNs in the context of skeleton-based action recognition, enabling the effective modeling of human joint correlations. The pioneer, STGCN\cite{yan2018spatial} predefines the physical structure of the human body as a fixed topology. 2s-AGCN\cite{shi2019two} introduces adaptiveness to topology learning with a self-attention mechanism. Since then, numerous studies\cite{ye2020dynamic,chen2021channel,shi2020skeleton,chi2022infogcn,lee2023hierarchically,xu2023language,zhou2024blockgcn} have aimed to capture the intrinsic correlations between human joints using learnable adjacency matrices, attention mechanisms or other techniques. Among them, AS-GCN\cite{li2019actional} and MS-G3D\cite{liu2020disentangling} leverage adjacency powering for multi-scale modeling, which is also adopted in our methods to establish topological symmetry constraints.

A previous work, DEGCN\cite{myung2024degcn}, introduces deformable temporal convolution for skeleton-based action recognition. However, DEGCN applies a uniform offset to all frames in a data-independent manner, resulting in the same receptive field across different frames. In contrast, our TSE-GCN learns a unique offset for each individual frame, which is data-dependent and enables more flexible receptive fields.

 \begin{figure*}[t]
	\centering
	%\fbox{\rule{0pt}{2in} \rule{0.9\linewidth}{0pt}}
	\includegraphics[width=0.9\linewidth]{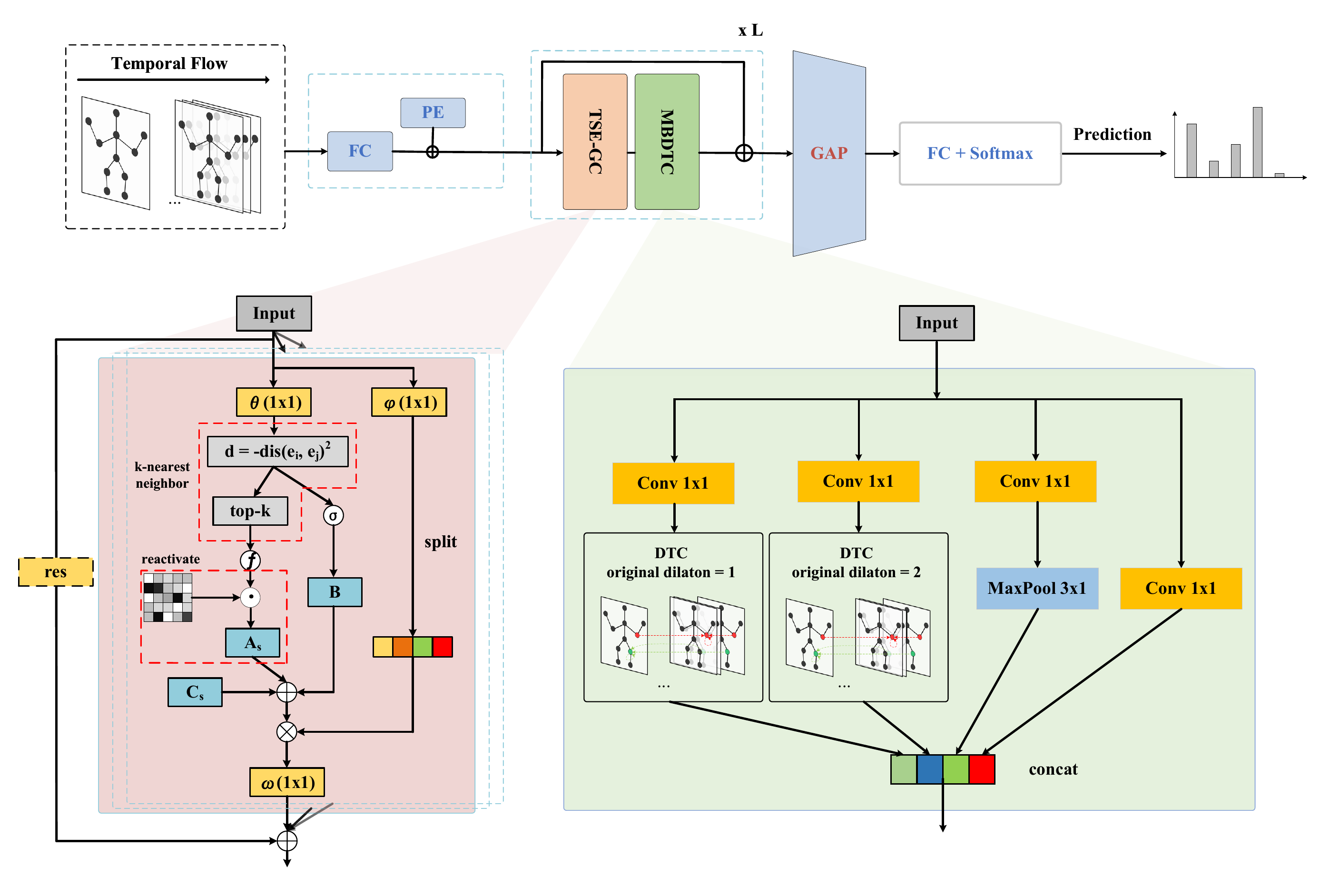}
	
	\caption{Architecture overview of the proposed TSE-GCN. PE denotes the learnable absolute positional embedding\cite{chi2022infogcn}. L denotes the number of stacked layers. In TSE-GC module, $f$ represents the generation function of scale mask where the indexes of top-k neighbors are mapped to relevant scales. $\otimes$, $\odot$, $\oplus$ denote matrix multiplication, element-wise multiplication, element-wise sum, respectively. Best viewed in color.}
	\label{fig:overall}
\end{figure*}

\subsection{Deformable Modeling}

The concept of deformable modeling originates from CNNs, with the main goal of directing important locations on images. As CNNs employ fixed convolutional filters, they struggle to adapt to complex and irregular shapes in images. Deformable convolution\cite{dai2017deformable} addresses this limitation by allowing the convolutional kernels to learn spatial offsets, enabling them to sample from non-grid locations in the input feature map. Deformable-DETR\cite{zhu2020deformable} builds upon the original DETR\cite{carion2020end} architecture with deformable attention mechanisms. Our MBDTC applies 1D temporal deformable convolution modules with multiple distinct origin sampling positions, along with several modifications to better suit the skeleton-based action recognition scenario.

\section{Method}
In this section, we first define related notations and revisit the formulations for vanilla graph convolution and its variants. Then we elaborate the design of our Topological Symmetry Enhanced Graph Convolution (TSE-GC) in Section \ref{section:TSE-GC} and Multi-Branch Deformable Temporal Convolution (MBDTC) in Section \ref{section:MBDTC}. The overall architecture of TSE-GCN is presented in the end.

\subsection{Preliminaries}
\textbf{Notations.} In GCNs, the human body within a motion sequence is represented as a spatio-temporal graph. The graph is denoted as $\mathcal{G} = (\mathcal{V}, \mathcal{E})$, where $\mathcal{V} = \{ v_1, v_2, \dots, v_N \}$ is the set of $N$ vertices representing joints and $\mathcal{E}$ is the edge set representing the correlations between joints. Typically, $\mathcal{G}$ is formulated by $\mathbf{X}\in\mathbb{R}^{N \times T \times C}$ and $\mathbf{A}\in\mathbb{R}^{N \times N}$. $\mathbf{X}$ is the feature tensor of $N$ vertices across $T$ frames, and $v_i$'s feature at frame $t$ is denoted as $x_{i,t}\in\mathbb{R}^{C}$. $C$ is the number of channels. $\mathbf{A}$ is the adjacency matrix, with its elements $a_{ij}$ representing the correlation between $v_{i,:}$ and $v_{j,:}$. 

\noindent \textbf{Formulations of Graph Convolution.} Within the realm of skeleton-based action recognition, the vanilla graph convolution proposed by \cite{kipf2016semi} is widely adopted:
\begin{equation}
	  \mathbf{X}_t^{(l+1)}=\sigma(\hat{\mathbf{A}}\mathbf{X}_t^{(l)}\mathbf{W}^{(l)}),
	  \label{eq:vanillagc}
\end{equation}
where $\mathbf{X}_t^{(l)}\in\mathbb{R}^{N \times d^{(l)}}$ denotes hidden feature representation of the $l$-th layer at frame $t$, $\hat{\mathbf{A}} = \mathbf{D}^{-\frac12}(\mathbf{A}+\mathbf{I})\mathbf{D}^{-\frac12}$ is the normalized adjacency matrix, $\mathbf{D}$ is the diagonal degree of $\mathbf{A}+\mathbf{I}$, $\mathbf{W}^{(l)}\in\mathbb{R}^{d^{(l)} \times d^{(l + 1)}}$ is learnable parameters of the $l$-th layer, and $\sigma$ indicates activation function. 

Variants of Eq. \ref{eq:vanillagc} typically adopt a partitioning strategy with multiple subsets involved. For example, ST-GCN\cite{yan2018spatial} divides $\mathbf{A} + \mathbf{I}$ into self-loop, centrifugal and centripetal components, Info-GCN\cite{chi2022infogcn} utilizes a multi-head self-attention mechanism where each head corresponds to a subset, and AS-GCN\cite{li2019actional} employs a multi-scale aggregation with each subset representing a distinct scale. The unified form can be formulated as:
\begin{equation}
	  \mathbf{X}_t^{(l+1)}=\sigma(\sum_{S} \mathrm{f}_s(\hat{\mathbf{A}}) \mathrm{g}_s(\mathbf{X}_t^{(l)})\mathbf{W}^{(l)}_s),
	  \label{eq:groupgc}
\end{equation}
where $\mathrm{f}_s(\cdot)$ and $\mathrm{g}_s(\cdot)$ represent mapping functions for adjacency matrices and feature tensors, respectively.

\subsection{Topology Learning with Symmetry Awareness}
\label{section:TSE-GC}
Topological symmetry is an inherent characteristic of the human body, where corresponding limbs mirror each other. Many physical activities, such as walking, running, or dancing, exhibit symmetrical patterns that should be captured through the process of graph representation learning. However, previous works tend to employ complex graph learning mechanisms without explicitly incorporating  topological symmetry during topology learning. 

On the other hand, most prior approaches typically utilize learnable adjacency matrices to adaptively capture correlations between joints, allowing the topology to be learned without any constraints based on physical bone connections. In contrast, directly using the physical structure as the topology for graph feature aggregation, as done in \cite{yan2018spatial}, yields suboptimal performance. Intuitively, a balanced approach that incorporates both flexibility and physical constraints is worth exploring.

Based on the analysis above, we argue that integrating topological symmetry constraints into topology learning could enhance the modeling capacity of graph convolution, which leads to TSE-GC. The architecture of our TSE-GC is illustrated in the bottom left corner of Fig.\ref{fig:overall}, and the forward process can be divided into two pathways, one for topology learning and one for feature partitioning. 

\noindent \textbf{Feature partitioning.} Given the input feature $\mathbf{X}\in\mathbb{R}^{N \times C}$, we first transform it into high-level representation with a linear embedding function $\psi$, which is formulated as:
\begin{equation}
	  \tilde{\mathbf{X}} = \psi(\mathbf{X}) = \mathbf{X}\mathbf{W}_{\psi},
	\label{eq:transform}
\end{equation}
where $\mathbf{W}_{\psi}\in\mathbb{R}^{C \times (C'K)}$ is the weight matrix and $\tilde{\mathbf{X}}\in\mathbb{R}^{N \times (C'K)}$ is the transformed feature. $\tilde{\mathbf{X}}$ is further divided along the channel dimension $C'K$ into $K$ partitions. 

The entire feature partitioning process, denoted as $\mathrm{g}_s(\cdot)$ in Eq. \ref{eq:groupgc}, is formulated as:
\begin{equation}
	  \tilde{\mathbf{X}} = \mathrm{g}_s(\mathbf{X}) = \psi_s(\mathbf{X}) = [\tilde{\mathbf{X}}_0 || \tilde{\mathbf{X}}_1 || ... || \tilde{\mathbf{X}}_{K-1}]
	\label{eq:partition}
\end{equation}
\begin{equation}
	  \tilde{\mathbf{X}}_k = \tilde{\mathbf{X}}_{:, kC': (k+1)C'},
	\label{eq:single_partition}
\end{equation}
where $||$ is concatenate function and $\tilde{\mathbf{X}}_k\in\mathbb{R}^{N \times C'}$ represents the $k$-th partition of the transformed feature $\tilde{\mathbf{X}}$. Note that $\psi_s(\cdot)$ represents distinct instances of $\psi(\cdot)$ applied to different subsets. 

\noindent \textbf{Topology learning.} The topology learning pathway is detailed in the left section of the illustrated TSE-GC shown in Fig.\ref{fig:overall}. A shared topology $\mathbf{M}\in\mathbb{R}^{N \times N}$ is utilized for all subsets and reactivated as $\mathbf{A}\in\mathbb{R}^{K\times N \times N}$. The reactivation could be interpreted as a form of resampling, where the correlations between joints at the relevant scales or hops are sampled. We use the term \textit{reactivate} to describe the characteristic of constraining topology learning in a symmetry manner, because the reactivated interactions can be updated through the training phase, while the other interactions remain unchanged. During the inference phase, only the reactivated interactions are utilized for graph feature aggregation. To maintain flexibility of topology modeling, we also introduce a data-dependent calibration matrix $\mathbf{B}\in\mathbb{R}^{N \times N}$, as well as a learnable adjacency matrix $\mathbf{C}\in\mathbb{R}^{K \times N \times N}$ for each subset. 

Specifically, we first employ another linear embedding function $\theta$ to embed the input feature for the subsequent topology learning, as in Eq. \ref{eq:transform}. The negative square of Euclidean distance $\mathbf{D}\in\mathbb{R}^{N \times N}$ is then calculated based on the embeddings between distinct joints, as $d_{ij} = -{dis(e_i, e_j)}^2$, where $e_i$ and $e_j$ represent the embeddings of the joints. $\mathbf{D}$ is further utilized to obtain the calibration matrix $\mathbf{B}$:
\begin{equation}
	  \mathbf{B} = \xi(\mathbf{D}),
	\label{eq:topologyB}
\end{equation}
where $\xi(\cdot)$ is the activation function. 

Based on the calculated Euclidean distance, the indexes of k-nearest neighbors can be identified as $\mathbf{IN}\in\mathbb{R}^{N \times K}$, which is mapped to the relevant scales or hops. Relevant scales are marked as $1$ in the scale mask. The generation of scale mask can be formulated as:
\begin{equation}
	  \mathbf{H} = f(\mathbf{IN}, \mathbf{D}_{sp}) = f(\mathbf{KNN}(\theta(\mathbf{X}), K), \mathbf{D}_{sp}),
	\label{eq:mapping}
\end{equation}
where $\mathbf{H}\in\mathbb{R}^{K \times N \times N}$ is the scale mask, $f$ is the generation function of scale mask where the indexes of top-k neighbors are mapped to relevant scales, $\mathbf{D}_{sp}$ is a mapping table based on the shortest path distance (SPD). The calculation of SPD is based on the physical connections, which can be formulated as:
\begin{equation}
        d_{sp_{i,j}} = \min_{P \in Paths(\mathcal{G})} \{|P|, P_1 = v_i, P_{|P|} = v_j\},
	\label{eq:SPD}
\end{equation}
where $d_{sp_{i,j}}$ represents SPD between nodes $v_i$ and $v_j$, and $P$ denotes the shortest path  in the graph $\mathcal{G}$ connecting these nodes. In practice, we calculate it with adjacency powering.

With the scale mask, we reactivate the shared topology to introduce topological symmetry constraints:
\begin{equation}
        \mathbf{A}_s = \mathbf{H}_s \odot \mathbf{M},
	\label{eq:reactivate}
\end{equation}
where $\odot$ is element-wise multiplication operation, $\mathbf{A}_s\in\mathbb{R}^{K \times N \times N}$ is the reactivated topology. We use subscript $s$ to indicate distinct scale masks and activated topology for different subsets.

Eventually, the learned topology for graph feature aggregation in our TSE-GC module is formulated as:
\begin{equation}
	  \mathbf{Z} = \mathrm{f}_s(\mathbf{A}) = \alpha \cdot \mathbf{A}_s + \beta \cdot \mathbf{B} + \mathbf{C}_s
	\label{eq:topology}
\end{equation}
\begin{equation}
	  \mathbf{Z}_k = \mathbf{Z}_{k,:,:},
	\label{eq:topology}
\end{equation}
where $\alpha$ and $\beta$ are trainable scalars that control the balance between flexibility and physical constraints. $\mathbf{Z}_k\in\mathbb{R}^{N \times N}$ represents the $k$-th partition of the learned topology, which is obtained by combining the activated topology, the learnable compensation term and the calibration term.

With Eq. \ref{eq:partition} and Eq. \ref{eq:topology} substituted into Eq. \ref{eq:groupgc}, we have our final formula for TSE-GC:
\begin{equation}
	\mathbf{X}_t^{(l+1)}=\sigma(\sum_{S}[\mathbf{Z}^{(l)}_0\tilde{\mathbf{X}}^{(l)}_0 ||  ... || \mathbf{Z}^{(l)}_{K-1}\tilde{\mathbf{X}}^{(l)}_{K-1}]\mathbf{W}^{(l)}_s),
	\label{eq:TSE-GC}
\end{equation}
 where $\sigma(\cdot)$ is the activation function. Some subscripts are simplified for readability. In general, our TSE-GC introduces topological symmetry constraints for the learning of $\mathbf{A}_s$ while maintaining a certain level of flexibility with $\mathbf{C}_s$, leading to an effective combination of prior-knowledge-based constraints and adaptability for topology learning. 

 \begin{figure}[t]
	\centering
	%\fbox{\rule{0pt}{2in} \rule{0.9\linewidth}{0pt}}
	\includegraphics[width=\linewidth]{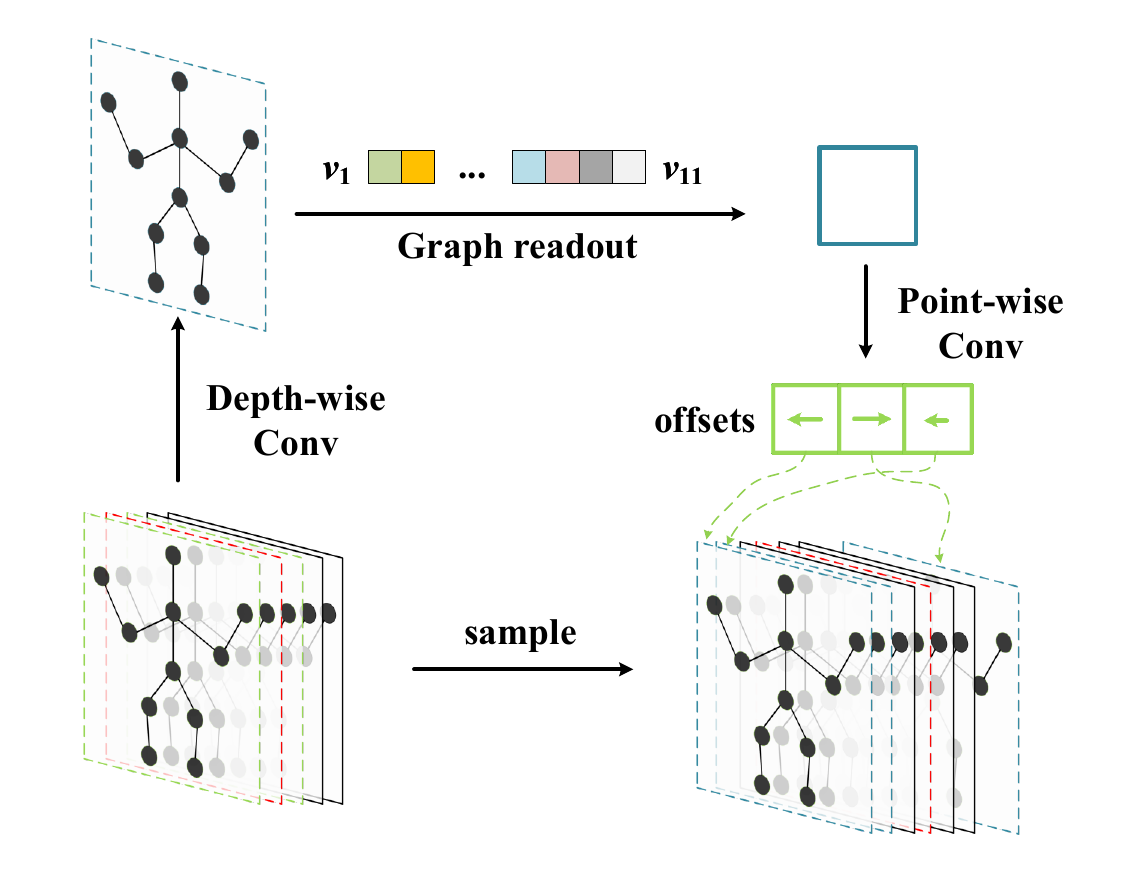}
	
	\caption{Sampling mechanism in our DTC Module. The receptive fields for the frame enclosed by the red dashed line are enclosed by the blue dashed line with the offsets. We assume the offsets to be integers for clarity. Best viewed in color.}
	\label{fig:DTC}
\end{figure}
\subsection{Multi-Branch Deformable Temporal Modeling}
\label{section:MBDTC}

Temporal modeling is essential for capturing complex motion dynamics in human action recognition. For GCNs, an effective temporal modeling module should be capable of flexibly capturing temporal dependencies, or correlations within the temporal graph. Previous works\cite{chen2021channel,liu2020disentangling,zhou2024blockgcn} typically utilize multi-scale temporal convolution to effectively capture both short-range and long-range dependencies with multiple fixed receptive fields, which is suboptimal. An intuitive solution is to introduce attention mechanisms, but their computational cost can be excessively high for long sequences. Therefore, we adopt the idea of introducing deformable modeling and construct our MBDTC with flexible receptive fields for skeleton-based action recognition, enabling dynamic temporal modeling. 

To begin with, we revisit the deformable convolution from \cite{dai2017deformable}. Given a 2D convolution, the sampling locations are defined by a regular grid $\mathcal{R}$, resulting in a fixed receptive field. The deformable convolution adds learnable offsets to make the sampling locations irregular, formulated as: 
\begin{equation}
    \mathbf{y}(\mathbf{p}_0) = \sum_{\mathbf{p}_n \in \mathcal{R}} \mathbf{w}(\mathbf{p}_n) \cdot \mathbf{x}(\mathbf{p}_0 + \mathbf{p}_n + \Delta \mathbf{p}_n),
    \label{eq:deformable2D}
\end{equation}
where $\mathbf{p}_0$ is a location on the output feature map and $\mathbf{p}_n$ is the fixed offsets in $\mathcal{R}$. $\Delta \mathbf{p}_n$ represents the learnable offset at each sampling point $\mathbf{p}_0 + \mathbf{p}_n$, which allows the convolution to adaptively adjust the receptive field.

For spatio-temporal graphs in GCNs, spatial features are inherently structured, while temporal features are discretely sampled from continuous frames. Typically, only correlations among the same joints across different frames are considered. Based on the distinct characteristic, we redesign the vanilla deformable convolution into our deformable temporal convolution (DTC) which learns the offsets for each frame with spatial graph feature extracted through weighted graph pooling. An example of our sampling mechanism is illustrated in Fig.\ref{fig:DTC}, where we insert a readout operation with graph pooling between depth-wise separable convolution.

Specifically, we first embed the input feature $\mathbf{X}\in\mathbb{R}^{N \times T \times C}$ into $\mathbf{Y}\in\mathbb{R}^{N \times T' \times C}$ with a depth-wise convolution. The potential reduction in frames from $T$ to $T'$ is caused by the stride applied to our DTC. The spatial graph feature of the embedding is then pooled, formulated as:
\begin{equation}
    \mathbf{Z} = \frac{1}{N} \sum_{n=1}^{N} w_n \mathbf{Y}_n,
    \label{eq:weightedpool}
\end{equation}
where $w_n$ is the weight scalar for joint $v_n$ and $\mathbf{Z}\in\mathbb{R}^{T' \times C}$ represents the graph readouts for $T'$ frames. Subsequently, the offsets are obtained with learnable parameters and $\mathbf{Z}$, formulated as:
\begin{equation}
    \mathbf{P} = \mathbf{Z} \mathbf{W},
    \label{eq:offsets}
\end{equation}
where $\mathbf{P}\in\mathbb{R}^{T' \times R}$ indicates the offsets for $T'$ frames. $R$ is the kernel size applied to our DTC. Notably, $R$ is also applied to the preceding depth-wise convolution. 

With the offsets and the original sampling locations, we adopt the linear interpolation for sampling and introduce a weight matrix $\mathbf{M}\in\mathbb{R}^{T' \times R}$, which is obtained from the offsets
\begin{table*}[t]
\centering
    \caption{Action classification performance on NTU RGB+D 60, NTU RGB+D 120 and NW-UCLA. \protect\dag indicates methods that are not directly comparable with further details in Section \protect\ref{csat}. * indicates the parameters or FLOPs are recalculated from their publicly available codes for a single stream in NTU RGB+D 120, then scaled by the number of streams for a fair comparison. The FLOPS are calculated using the fvcore library\protect\footnotemark. We omit the results without publicly available codes for reproduction.}
    \label{tab:all}
\resizebox{1.0\textwidth}{!}{
\begin{tabular}{lllllccccc}

\toprule
\multirow{2}{*}{Methods} & \multirow{2}{*}{Publication} & \multirow{2}{*}{Modalities} & \multirow{2}{*}{Params(M)} & \multirow{2}{*}{FLOPs(G)} & \multicolumn{2}{c}{NTU RGB+D 60} & \multicolumn{2}{c}{NTU RGB+D 120} & \multirow{2}{*}{NW-UCLA}\\
&  & & & & X-Sub(\%) & X-View(\%) & X-Sub(\%) & X-Set(\%) \\ \hline \hline
\addlinespace
ST-GCN\cite{yan2018spatial}& AAAI 2018& J& 3.1\textsuperscript{*}& 3.48\textsuperscript{*}& 81.5& 88.3& 70.7& 73.2& - \\
2S-AGCN\cite{shi2019two}& CVPR 2019& J+B& 6.8\textsuperscript{*}& 7.96\textsuperscript{*}& 88.5& 95.1& 82.5& 84.2& - \\
DC-GCN+ADG\cite{cheng2020decoupling}& ECCV 2020& J+B+JM+BM& 19.6& 7.32& 90.8& 96.6& 86.5& 88.1& 95.3\\
Shift-GCN\cite{cheng2020skeleton}& CVPR 2020& J+B+JM+BM& -& -& 90.7& 96.5& 85.9& 87.6& 94.6\\
MS-G3D\cite{liu2020disentangling}& CVPR 2020& J+B& 12.8\textsuperscript{*}& 28.24\textsuperscript{*}& 91.5& 96.2& 86.9& 88.4& - \\
MST-GCN\cite{chen2021multi}& AAAI 2021& J+B+JM+BM& 12.0& -& 91.5& 96.6& 87.5& 88.8& - \\ 
CTR-GCN\cite{chen2021channel}& ICCV 2021& J+B+JM+BM& 5.6\textsuperscript{*}& 7.88\textsuperscript{*}& 92.4& 96.4& 88.9& 90.4& 96.5\\
EfficientGCN-B4\cite{song2022constructing}& TPAMI 2022& J+B+JM+BM& 8.0 & - & 91.7& 95.7& 88.3& 89.1& -\\
Info-GCN\cite{chi2022infogcn}& CVPR 2022& J+B+JM+BM& 6.3& 6.72& 92.3& 96.7& 89.2& 90.7& 96.6\\
FR-Head\cite{zhou2023learning}& CVPR 2023& J+B+JM+BM& 8.0& -& 92.8& 96.8& 89.5& 90.9& 96.8\\
BlockGCN\cite{zhou2024blockgcn}& CVPR 2024& J+B+JM+BM& 5.2\textsuperscript{*}& 8.20\textsuperscript{*}& 93.1& 97.0& 90.3& 91.5& 96.9\\
\midrule
Info-GCN\textsuperscript{\dag}\cite{chi2022infogcn}& CVPR 2022& -& 9.4& 10.08& 93.0& 97.1& 89.8& 91.2& 97.0\\
HD-GCN\textsuperscript{\dag}\cite{lee2023hierarchically}& ICCV 2023& -& 10.1& 9.60& 93.4& 97.2& 90.1& 91.6& 97.2\\
Shap-Mix\textsuperscript{\dag}\cite{zhang2024shap}& CVPR 2024& J+B+JM+BM& -& -& 93.7& 97.1& 90.4& 91.7& - \\
\midrule
\textbf{TSE-GCN(Ours)}& & J& \textbf{1.1}& \textbf{1.38}& 90.8& 95.3& 86.6& 88.2& 95.3\\
\textbf{TSE-GCN(Ours)}& & J+B+JM+BM& \textbf{4.4}& \textbf{5.52}& 92.9& 96.8& 90.0& 91.1& 96.9\\
\bottomrule
\end{tabular}
}
\end{table*}
through a linear transformation. Our DTC can be formulated as:
\begin{equation}
    \mathbf{X}^{(l+1)} = \sum_{r=1}^{R} \mathbf{W}_R \Gamma(\mathbf{P}, \mathbf{X}^{(l+1)}, \mathbf{M}),
    \label{eq:DTC}
\end{equation}
where $\Gamma(\cdot,\cdot,\cdot)$ is the sampling and reweighting function, $\mathbf{W}_R$ is learnable parameters and $\mathbf{X}^{(l+1)}\in\mathbb{R}^{N \times T' \times C}$ is the output with aggregated temporal features. 

With our DTC module, we construct MBDTC module following the multi-scale temporal convolution module (MS-TCN) adopted by \cite{chen2021channel}. The architecture of our MBDTC is illustrated in the bottom right corner of Fig.\ref{fig:overall}, which is consists of four branches, with each branch containing a $1 \times 1$ convolution to reduce channel dimension. The first two branches utilize our DTC with different original sampling locations defined by different dilations, the other two branches implement a max-pooling operation and an identity function, respectively. The outputs are concatenated as the final output.  

\subsection{Model Architecture}
Based on the aforementioned TSE-GC and MBDTC, we construct a topological symmetry enhanced graph convolutional network TSE-GCN for skeleton-based action recognition. The model architecture is illustrated in Fig.\ref{fig:overall}. Based on experimental results, we select GeLU as the activation function in the network, except for Eq. \ref{eq:topologyB} where we use Tanh for a more suitable output range in modeling topology. 

The input feature is first embedded with a linear transformation and then combined with a learnable absolute positional embedding from \cite{chi2022infogcn}. Subsequently, the embeddings are fed into a stack of our basic blocks, each consisting of a TSE-GC module, a MBDTC module and residual connections. $L=9$, is the number of times our basic block is stacked. The number of channels for nine blocks are 64-64-64-64-128-128-128-256-256. Following the stacked basic blocks, a global average pooling operation and a softmax classifier is applied to generate predictions across different action classes.

\section{Experiments}

% \footnotetext{\url{https://github.com/facebookresearch/fvcore}}
\subsection{Datasets}
\noindent \textbf{NTU RGB+D.} NTU RGB+D\cite{shahroudy2016ntu} is a large-scale human action recognition dataset containing 56,880 skeleton action sequences over 60 classes. The action samples are performed by 40 participants in different age groups, and each sample contains an action and is guaranteed to have at most 2 subjects. For each sample, the skeleton data of 25 joints is captured by three Microsoft Kinect v2 cameras from different views concurrently. Two benchmarks are provided by the authors: (1) cross-subject (X-sub): training data comes from 20 subjects, and testing data comes from the other 20 subjects. (2) cross-view (X-view): data captured from the front and two side views is used for training, and data captured from the left and right 45-degree views is used for testing. 

\vspace{1em}
\noindent \textbf{NTU RGB+D 120.} NTU RGB+D 120\cite{liu2019ntu} is an extended version of NTU RGB+D with additional 57,367 skeleton sequences over 60 extra action classes, which is one of the largest datasets with 3D joints annotations for human action recognition. Totally 113,945 samples over 120 classes are performed by 106 volunteers with 32 distinct setups for locations and backgrounds, captured with three cameras views. The authors recommend two benchmarks: (1) cross-subject (X-sub): training data comes from 53 subjects, and testing data comes from the other 53 subjects. (2) cross-setup (X-set): training data comes from samples with even setup IDs, and testing data comes from samples with odd setup IDs.

\vspace{1em}
\noindent \textbf{NW-UCLA.} NW-UCLA\cite{wang2014cross} is a small human action recognition dataset containing 1494 video clips over 10 action categories. Each action is performed by 10 subjects, with the skeleton data captured by three Kinect cameras simultaneously from multiple viewpoints. Following the evaluation protocol recommended by the authors, we use the viewpoints of the first two cameras for training and the other for testing.

\subsection{Implementation Details}
All experiments are conducted on two RTX 3090 GPUs with the PyTorch deep learning framework. The SGD optimizer is employed with a Nestrov momentum of 0.9 and a weight deacy of 0.004 for NTU RGB+D and NTU RGB+D 120, and 0.002 for NW-UCLA. Our method utilizes cross-entropy loss. The learning rate is initialized at 0.05 and reduced by a factor of 0.1 at epoch 110 and 120, with a total epoch 140. For NTU datasets, we set a batch size of 64 with each sample resized to 64 frames and adopt the data-processing in \cite{zhang2020semantics}. For NW-UCLA, the batch size is selected as 16 and we adopt he data-processing in \cite{cheng2020skeleton}.

\subsection{Comparison with the State-of-the-art}
\label{csat}
In this subsection, we compare the proposed TSE-GCN against state-of-the-art methods on NTU RGB+D, NTU RGB+D 120 and NW-UCLA in Table \ref{tab:all}. To establish a fair comparison, we follow the commonly adopted four-stream fusion approach in our experiment. Specifically, we fuse the results of \textit{joint}, \textit{bone}, \textit{joint motion} and \textit{bone motion} modalities. Notably, some of the methods are not directly comparable to our methods, which are shown in the second part of the table, for example, HD-GCN\cite{lee2023hierarchically} applies a six-stream ensemble with their hand-crafted modalities, as well as Info-GCN\cite{chi2022infogcn}, and Shap-Mix\cite{zhang2024shap} utilizes the additional mixed data from their data augmentation approach which doubles the number of samples. The performance of Info-GCN in the first part is from the reproduction from \cite{huang2023graph} for a fair comparison, following \cite{zhou2024blockgcn}. 

Experimental results demonstrate that our method, TSE-GCN achieves state-of-the-art performance on the common benchmarks. Compared with the state-of-the-art BlockGCN\cite{zhou2024blockgcn}, the proposed TSE-GCN achieves 0.3\% lower accuracies on average with 15.4\% fewer parameters and 32.7\% fewer FLOPs. Compared with FR-Head\cite{zhou2023learning}, the proposed TSE-GCN surpasses it with 45\% fewer parameters. It is noteworthy that we prioritize a better trade-off between model size and performance in our hyper-parameter setting instead of accuracies, as detailed in Section \ref{sec:ablation}. Meanwhile, our method can serve as a backbone and be combined with Shap-Mix\cite{zhang2024shap} or FR-Head\cite{zhou2023learning} to further improve performance.

\subsection{Ablation Study}
\label{sec:ablation}
In this subsection, we analyze the proposed TSE-GC and its configurations along with MBTDC on the X-sub benchmark of the NTU RGB+D 120 dataset, using joint stream data. We reconstruct ST-GCN\cite{yan2018spatial} with our selection of activation function as the baseline in our experiments, with the $2$-nd layer removed.

% \begin{table}[h]
% \centering
% \caption{Comparison of performance when adding PE and modules of our TSE-GC gradually. TSE-GC is examined through distinct topologies to assess their effectiveness. PE denotes the learnable absolute positional embedding \cite{chi2022infogcn} added to the input feature before the first layer of our network.}

% \label{tab:dtc}
% \resizebox{0.5\textwidth}{!}{
% \begin{tabular}{ccc}
% \toprule
% Methods& Params & Acc(\%) \\
% \midrule
% Baseline &  3.0M & 85.0 \\
% \indent w MS-TCN&  3.1M &  \\
% \checkmark & 1.2M(-1.9M) & 85.4 \\
% \bottomrule
% \end{tabular}
% }
% \end{table}

\begin{table}[h]
\centering
\caption{Comparisons of performance when adding PE and our TSE-GCN gradually. TSE-GC is examined through distinct topologies to assess their effectiveness. PE denotes the learnable absolute positional embedding \cite{chi2022infogcn} added to the input feature before the first layer of our network.}

\label{tab:ablation}
\resizebox{0.48\textwidth}{!}{
\begin{tabular}{ccccccc}
\toprule
\multirow{2}{*}{Baseline}&  \multirow{2}{*}{MBTDC}& \multirow{2}{*}{PE}& \multicolumn{2}{c}{TSE-GC} & \multirow{2}{*}{Params(M)} & \multirow{2}{*}{Acc(\%)} \\
&&&$\mathbf{C}_s$& $\mathbf{A}_s + \mathbf{B}$ &&\\
\midrule
\checkmark &  - & - & - & -& 3.0 & 84.9 \\
\checkmark &  \checkmark & - & - && 1.2(-1.9) & 85.4\\
\checkmark &  \checkmark &\checkmark& - & - & 1.2 & 85.5 \\
\checkmark &  \checkmark &\checkmark& \checkmark & - & 1.0 & 86.4 \\
\checkmark &  \checkmark &\checkmark& -& \checkmark & 1.1 & 86.2\\
\checkmark &  \checkmark &\checkmark& \checkmark & \checkmark & 1.1 & 86.6 \\
\bottomrule
\end{tabular}
}
\end{table}

\begin{table}[h]
\centering
\caption{Comparisons of our TSE-GCN with different settings. $K$ denotes the number of partitions of input feature and adjacency matrices. $R$ denotes the ratio of the embedded number of channels to the original.}

\label{tab:configuration}
\resizebox{0.3\textwidth}{!}{
\begin{tabular}{cccc}
\toprule
$K$ & $R$ & Params(M)& Acc(\%)\\
\midrule
2&  4 &  1.3 & 86.4\\
 &  8 &  1.0 & 86.5\\
\midrule
3&  4 & 1.7 & 86.7\\
 &  8 & 1.1 & 86.6\\
 &  16 & 0.9 & 86.5\\
\midrule
4&  8 & 1.3 & 86.5\\
5&  8 & 1.5 & 86.3\\
\bottomrule
\end{tabular}
}
\end{table}

\noindent \textbf{Effectiveness of MBDTC and TSE-GC.} We examine the effectiveness of each component of our TSE-GCN by adding them to the baseline gradually. The experimental results are shown in Table \ref{tab:ablation}. First, we replace the original TCN with our MBTDC, resulting in a 0.5\% improvement in accuracy while reducing the number of parameters by 1.9M. This enhancement demonstrates the effectiveness of MBTDC. The introduction of PE marginally increases the parameter count with a slight improvement of 0.1\% in accuracy.

We then validate the effect of our TSE-GC by decoupling it into separate learned topologies $\mathbf{C}_s$ and $\mathbf{A}_s + \mathbf{B}$. For example, $\mathbf{C}_s$ indicates a setting of our TSE-GC without the learning of $\mathbf{A}_s + \mathbf{B}$. $\mathbf{A}_s + \mathbf{B}$ is not further decoupled because the learning of $\mathbf{B}$ contributes to the gradients of the embedding function at the initial stages of topology learning pathway. The fully flexible setting of our TSE-GC ($\mathbf{C}_s$) improves the performance by 0.9\%, while the constrained setting ($\mathbf{A}_s + \mathbf{B}$) yields an improvement of 0.7\%. By integrating both, our TSE-GC outperform the baseline by 1.7\% while reducing 63\% of its parameters.

% ,  which corresponds to our analysis in Section \ref{section:TSE-GC}

\noindent \textbf{Configurations of TSE-GC.} We explore different configurations of our TSE-GC, specifically the number of partitions $K$ and the reduction rate $R$ of embedding function $\psi$ and $\theta$. Since $K$ also represents different scales within TSE-GC, we examine settings of $K={2,3,4,5}$ to analyze its impact on performance. When $K=1$, $\mathbf{A}_s$ falls into $\mathbf{I}$ as we do not exclude the joint itself in our KNN implementation to preserve self-connection. When $K=7$ and each neighboring joint represents a different scale, $\mathbf{A}_s$ functions as a fully connected layer on NTU datasets.

As shown in Table \ref{tab:configuration}, our TSE-GC consistently outperforms the baseline under all configurations, demonstrating its robustness to diverse settings. By comparing the experimental results of $K={2,3,4,5}$, we find our models achieve best results when $K=3$. We argue that fewer partitions lead to an insufficient modeling while more partitions introduce redundancy. As for the setting of $R$, a lower value of $R$ leads to more adjacency matrices and additional parameters, thereby increasing the overall computational complexity.
Our optimal performance can be achieved with $K=3$ and $R=4$. However, in order to balance performance and efficiency, we select $K=3$ and $R=8$ as the final choice of hyper parameters for TSE-GC.

\begin{figure}[t]
    \centering
    \begin{subfigure}[b]{0.22\textwidth}
        \centering
        \includegraphics[width=1.0\linewidth]{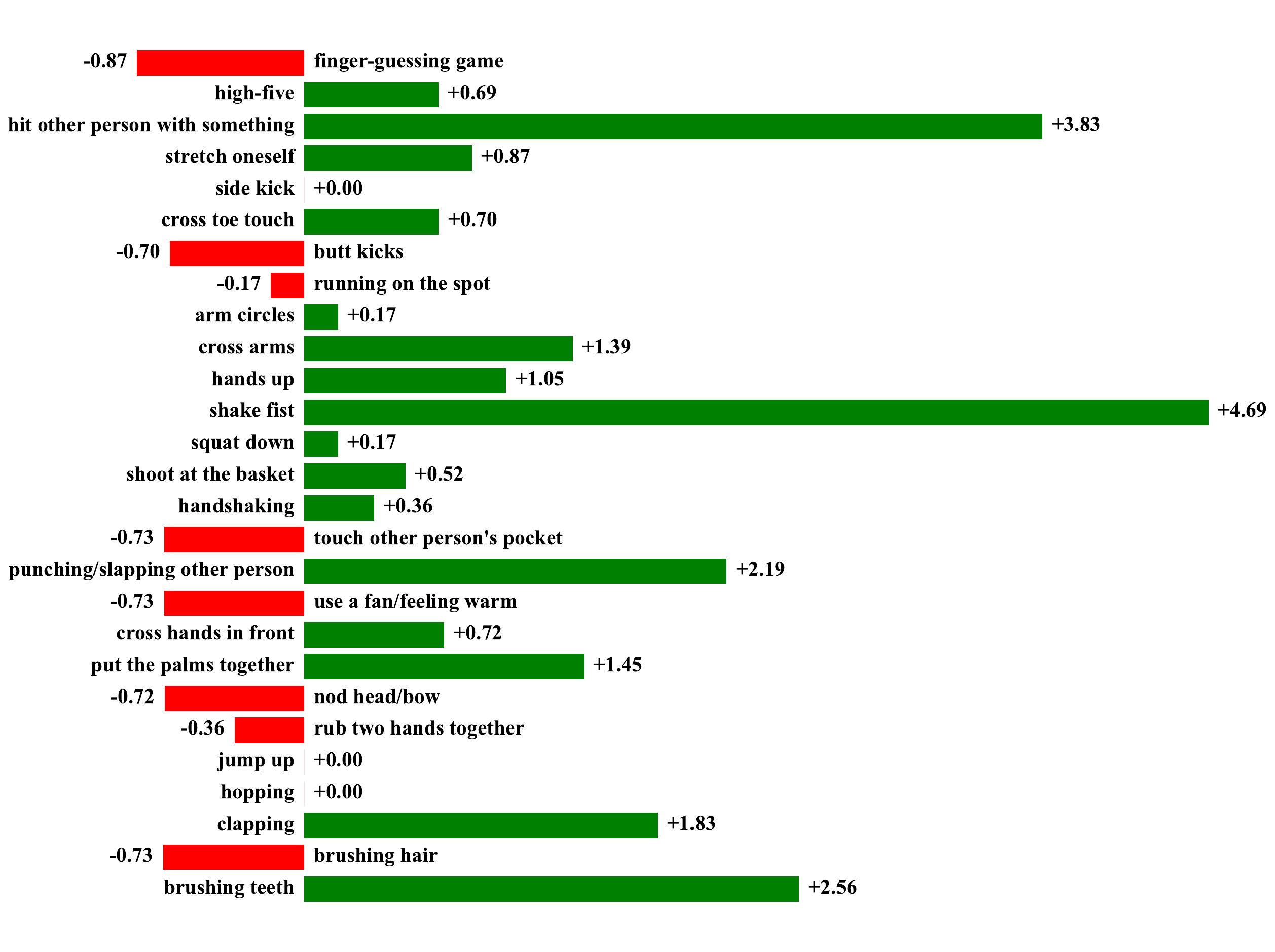}
        \caption{Comparison with ST-GC.}
        \label{fig:sub1}
    \end{subfigure}
    \hfill
    \begin{subfigure}[b]{0.22\textwidth}
        \centering
        \includegraphics[width=1.0\linewidth]{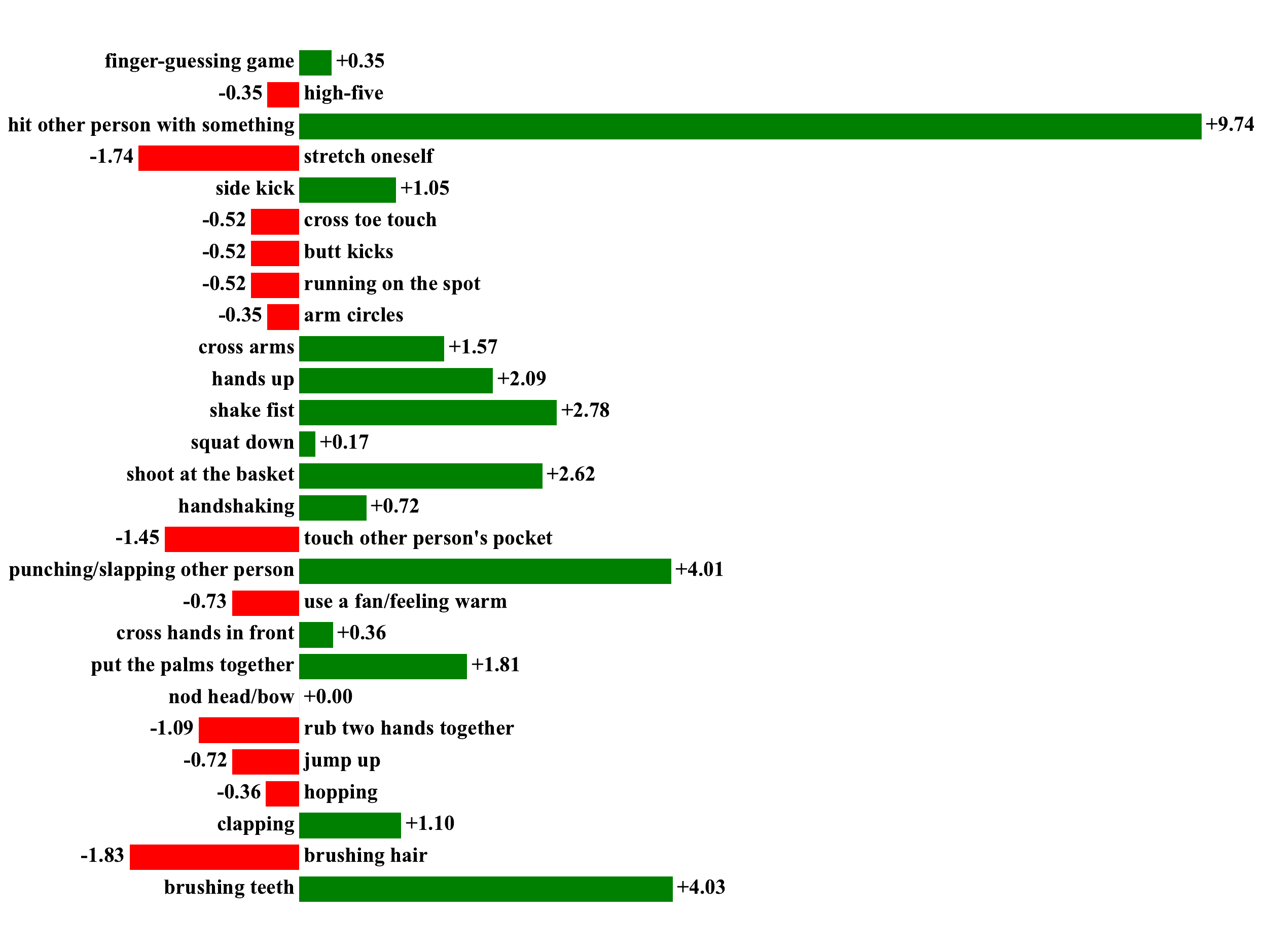}
        \caption{Comparison with CTR-GC.}
        \label{fig:sub2}
    \end{subfigure}
    \caption{Accuracy difference(\%) between TSE-GC and two representative GCs on symmetry related classes generated by GPT4 \cite{achiam2023gpt}. Green bars indicate improvements.}
    \label{fig:accuracies}
\end{figure}

\subsection{Performance Analysis on Single Stream}
In table \ref{tab:single}, we compare single-stream performance of our TSE-GCN with other GCNs. Our model achieves a competitive performance on NTU RGB+D 120 joint stream with the lowest computational complexity. 

\begin{table}[h]
\centering
\caption{Comparisons of GCNs on single stream.}
\label{tab:single}
\resizebox{0.48\textwidth}{!}{
\begin{tabular}{ccccc}
\toprule
Methods & X-Sub(\%) & X-Set(\%) & Params(M) & FLOPs(G)\\
\midrule
CTR-GCN & 84.9& 86.5& 1.5 & 1.97 \\
Info-GCN & 85.1& 86.3& 1.6 & 1.68 \\
HD-GCN & 85.7& 87.3 & 1.7 & 1.60 \\
FR-Head & 85.5 & 87.3 & 2.0 & - \\
BlockGCN & 86.9 & 88.2 & 1.3 & 2.05 \\
\midrule
TSE-GCN & 86.6 & 88.2 & 1.1 & 1.38 \\
\bottomrule
\end{tabular}
}
\end{table}
% 选择哪些节点需要考虑对称性

We further analyze the accuracies of certain classes to confirm that the experimental results align with the intended design to incorporate topological symmetry awareness. Specifically, we utilize GPT4\cite{achiam2023gpt} to identify symmetry related classes within NTU RGB+D 120, as the correlation between certain actions and topological symmetry is abstract and hard to be defined by human. The final list contains 27 classes, including clapping, hands up, cross arms and so on. We calculate the accuracy difference of these classes between TSE-GC and two representative GCs in Fig.\ref{fig:accuracies}. We select ST-GC and CTR-GC as the topology of ST-GC is fully physical, while the topology of CTR-GC is nearly fully flexible due to the catastrophic forgetting of skeletal topology\cite{zhou2024blockgcn}. Experimental results demonstrate that TSE-GC shows an average improvement of +0.6\% compared to ST-GC and +0.8\% compared to CTR-GC on these classes, which corresponds to their distinct topologies. Meanwhile, overall performance improvement is observed in both cases, with a few classes showing gains exceeding the average, such as +4.69\% for \enquote{shake fist} in ST-GC case and +9.74\% for \enquote{hit other person with something} in CTR-GC case.
\section{Conclusion}
In this work, we propose a novel topological symmetry enhanced graph convolution network (TSE-GCN) for skeleton-based action recognition. Our TSE-GC module leverages the inherent topological symmetry of the human body to achieve an effective and balanced topology learning between flexibility and physical constraints. Our multi-branch deformable temporal convolution (MBTDC) module incorporates deformable modeling to enable flexible receptive fields. TSE-GCN achieves competitive performance compared with state-of-the-art methods on three datasets, with the effectiveness of each component validated. Our future research may proceed in two main directions: 

% We also introduce the concept of deformable modeling and construct a multi-branch deformable temporal convolution (MBTDC) module for GCNs.
\noindent(1) Efficient cross-spacetime feature aggregation. With flexible receptive fields in deformable temporal convolution, it is possible to achieve a learnable cross-spacetime feature aggregation, leading to a unified and powerful representation for GCNs.  

\noindent(2) Redundancy reduction in graph convolution. Some actions can be identified with only topological symmetric joints, such as \enquote{clapping} and \enquote{hands up}. Removing redundant correlations and preserving informative ones could enhance efficiency and facilitate practical applications.

{
    \small
    \bibliographystyle{ieeenat_fullname}
    \bibliography{main}
}

% 正文的参考文献
% {
%     \small
%     \printbibliography[heading=bibintoc, title={References}] % 正文引用
% }

% WARNING: do not forget to delete the supplementary pages from your submission 
% \begin{refsection}
% \input{sec/X_suppl} % 附录内容
% % \clearpage 
% {
%     \small
%     \printbibliography[heading=subbibliography, title={References}]
% }
% \end{refsection}
% \input{sec/X_suppl}
% {
%     \small
%     \bibliographystyle{ieeenat_fullname} % 引用样式
%     \bibliography{appendix} % 附录引用使用 appendix.bib 文件
% }
\end{document}